\documentclass[letterpaper, 10 pt, conference]{ieeeconf}  %

\IEEEoverridecommandlockouts                              %

\usepackage{cite}

\makeatletter
\let\NAT@parse\undefined
\makeatother

\usepackage[utf8]{inputenc} %
\usepackage[T1]{fontenc}    %
\usepackage{hyperref}       %
\usepackage{url}            %
\usepackage{booktabs}       %
\usepackage{array,caption}
\usepackage{amsfonts}       %
\usepackage{nicefrac}       %
\usepackage{microtype}      %

\usepackage{amsmath}
\usepackage{amssymb}

\usepackage{enumerate}
\usepackage[usenames, dvipsnames]{color}

\usepackage{graphicx}
\usepackage{bm}

\usepackage{subfigure}
\usepackage{adjustbox}

\graphicspath{{plots/}}

\usepackage{makecell}

\usepackage{algorithm}
\usepackage[noend]{algpseudocode}

\algnewcommand\algorithmicinput{\textbf{Initialize:}}
\algnewcommand\INPUT{\item[\algorithmicinput]}

\usepackage{thm-restate}

\usepackage{wrapfig}
\usepackage{changepage}
\usepackage{float}
\usepackage{bm}

\usepackage{multirow}

\usepackage{relsize}
\DeclareMathOperator*{\argmin}{argmin}

\definecolor{mygray}{rgb}{0.4,0.4,0.4}

\title{\LARGE \bf
CW-ERM: Improving Autonomous Driving Planning with \\ Closed-loop Weighted Empirical Risk Minimization
}

\author{Eesha Kumar$^{1,*}$, Yiming Zhang$^{2}$, Stefano Pini$^{1}$, Simon Stent$^{1}$, \\Ana Ferreira$^{2}$, Sergey Zagoruyko$^{1}$, Christian S. Perone$^{1,*}$%
\thanks{$^{1}$Author is with Woven Planet United Kingdom Limited, 114-116 Curtain Road, London, United Kingdom, EC2A 3AH.
        {\tt\small {firstname}.{lastname}@woven-planet.global}}%
\thanks{$^{2}$Author is with Woven Planet North America, Inc., 900 Arastradero Rd, Palo Alto, CA, USA 94304.
        {\tt\small {firstname}.{lastname}@woven-planet.global}}%
\thanks{$^{*}$Equal contribution.}%
}
\begin{document}

\maketitle
\thispagestyle{empty}
\pagestyle{empty}

\begin{abstract}
The imitation learning of self-driving vehicle policies through behavioral cloning is often carried out in an open-loop fashion, ignoring the effect of actions to future states. Training such policies purely with Empirical Risk Minimization (ERM) can be detrimental to real-world performance, as it biases policy networks towards matching only open-loop behavior, showing poor results when evaluated in closed-loop.
In this work, we develop an efficient and simple-to-implement principle called Closed-loop Weighted Empirical Risk Minimization (CW-ERM), in which a closed-loop evaluation procedure is first used to identify training data samples that are important for practical driving performance and then we these samples to help debias the policy network. We evaluate CW-ERM in a challenging urban driving dataset and show that this procedure yields a significant reduction in collisions as well as other non-differentiable closed-loop metrics.
\end{abstract}

\section{Introduction}
Learning effective planning policies for self-driving vehicles (SDVs) from data such as human demonstrations remains one of the major challenges in robotics and machine learning. Since early works such as ALVINN~\cite{Pomerleau1989}, Imitation Learning has seen major recent developments using modern Deep Neural Networks (DNNs)~\cite{bansal2018chauffeurnet, end-to-end-2017,nvidia-end-to-end-2016,conditional-il-2018,gail-2017,vitelli2022safetynet}. Imitation Learning (IL), and especially Behavioral Cloning (BC), however, still face fundamental challenges~\cite{codevilla2019exploring}, including causal confusion~\cite{causal-confusion} (later identified as a feedback-driven covariate shift\cite{three-regimes}) and dataset biases~\cite{codevilla2019exploring}, to name a few.

There is one particular limitation of IL policies trained with BC that is, however, often overlooked: the mismatch between training and inference-time execution of the policy actions. Most of the time, BC policies are trained in an open-loop fashion, predicting the next action given the immediate previous action and optionally conditioned on recent past actions ~\cite{bansal2018chauffeurnet, end-to-end-2017,nvidia-end-to-end-2016,conditional-il-2018,vitelli2022safetynet}. These policies, however, when executed in real-world, impact the future states. Small prediction errors can drive covariate shift and make the network predict in an out-of-distribution regime.

In this work, we address the mismatch between training and inference through the development of a simple training principle. Using a closed-loop simulator, we first identify and then reweight samples that are important for the closed-loop performance of the planner. We call this approach \textbf{CW-ERM} (Closed-loop Weighted Empirical Risk Minimization), since we use Weighted ERM~\cite{covshift} to correct the training distribution in favour of closed-loop performance. We extensively evaluate this principle on real-world urban driving data and show that it can achieve significant improvements on planner metrics that matter for real-world performance (e.g. collisions).

Our contributions are therefore the following:
\begin{itemize}
    \item We motivate and propose Closed-loop Weighted Empirical Risk Minimization (CW-ERM), a technique that leverages closed-loop evaluation metrics acquired from policy rollouts in a simulator to debias the policy network and reduce the distributional differences between training (open-loop) and inference time (closed-loop);
    \item we evaluate CW-ERM experimentally on a challenging urban driving dataset in a closed-loop fashion to show that our method, although simple to implement, yields significant improvements in closed-loop performance without requiring complex and computationally expensive closed-loop training methods;
    \item we also show an important connection of our method to a family of methods that addresses covariate shift through density ratio estimation.
\end{itemize}
In Section~\ref{sec:methodology}, we detail the proposed CW-ERM and in Section~\ref{sec:experimental} we show the CW-ERM experiments and compare them against ERM.

\section{Methodology}
\label{sec:methodology}
\begin{figure*}[ht]
    \centering
\includegraphics[width=\textwidth]{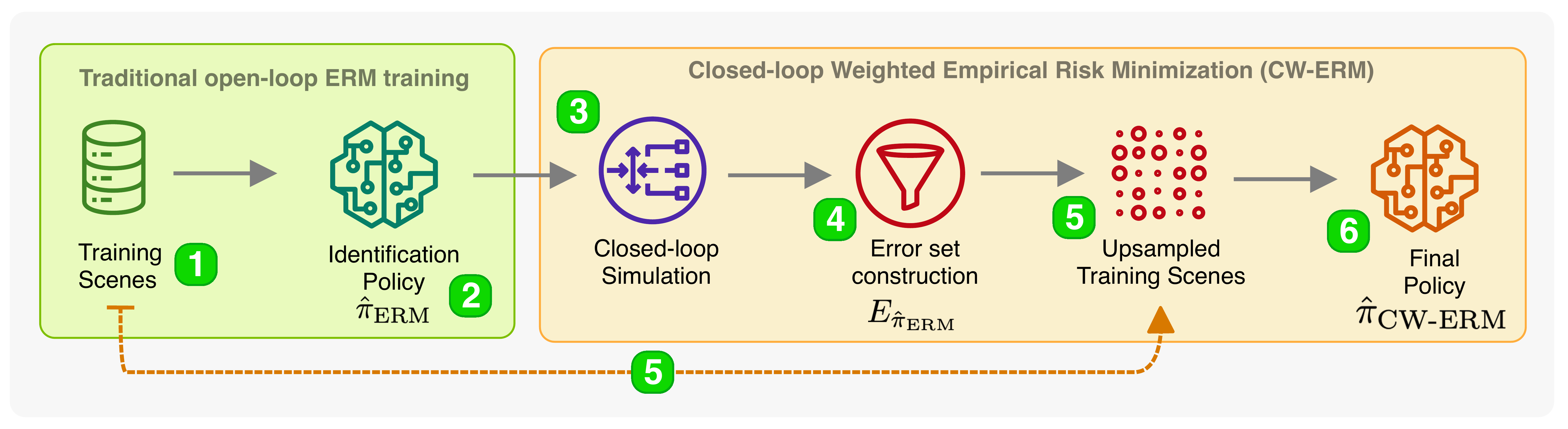}
    \caption{High-level overview of our proposed Closed-loop Weighted Empirical Risk Minimization (CW-ERM) method. In steps \textbf{(1-2)} we train an identification policy $\hat{\pi}_{\text{ERM}}$ using traditional ERM~\cite{vapnik1991} on a set of training data samples or driving "scenes". In step \textbf{(3)}, we perform closed-loop simulation of the policy $\hat{\pi}_{\text{ERM}}$ and collect metrics to construct the error set in step \textbf{(4)}. With the error set in hand, we upsample scenes in the training set as shown in step \textbf{(5)}. We train the final policy $\hat{\pi}_{\text{CW-ERM}}$ using CW-ERM as shown in step \textbf{(6)} with the upsampled $\mathcal{D}_{\text{up}}$ set.}
    \label{fig:high_level_arch}
\end{figure*}

\subsection{Problem Setup}
The traditional formulation of supervised learning for imitation learning, also called behavioral cloning (BC), can be formulated as finding the policy $\hat{\pi}_{BC}$:

\begin{equation}
\label{eqn:bc-erm}
\hat{\pi}_{BC} = \argmin_{\pi \in \Pi} \mathbb{E}_{s \sim d_{\pi^*}, a \sim \pi^*(s)}[\ell(s,a,\pi)]
\end{equation}

where the state $s$ is sampled from the expert state distribution $d_{\pi^*}$ induced when following the expert policy $\pi^*$. Actions $a$ are sampled from the expert policy $\pi^*(s)$. The loss $\ell$ is also known as the surrogate loss that will find the policy $\hat{\pi}_{BC}$ that best mimics the unknown expert policy $\pi^*(s)$. In practice, we only observe a finite set of state-action pairs ${(s_i, a^*_i)}_{i=1}^m$, so the optimization is only approximate and we then follow the Empirical Risk Minimization (ERM) principle to find the policy $\pi$ from the policy class $\Pi$.

If we let $\mathbb{E}_{s \sim d_{\pi^*}, a \sim \pi^*(s)}[\ell(s,a,\pi)] = \epsilon$, then it follows that $J(\pi) \leq J(\pi^*) + T^2 \epsilon$ as shown by the proof in~\cite{pmlr-v9-ross10a}, where $J$ is the total cost and $T$ is the task horizon. As we can see, the total cost can grow quadratically in $T$.

When the policy $\hat{\pi}_{BC}$ is deployed in the real-world, it will eventually make mistakes and then induce a state distribution $d_{\hat{\pi}_{BC}}$ different than the one it was trained on ( $d_{\pi^*}$). During closed-loop evaluation of driving policies, non-imitative metrics such as collisions and comfort are also evaluated. However, they are often ignored in the surrogate loss or only implicitly learned by imitating the expert due to the difficulty of overcoming differentiability requirements, as smooth approximations of these metrics are still different than the non-differentiable counterparts often used. These policies can often show good results in open-loop training, but perform poorly in closed-loop evaluation or when deployed in a real SDV due to the differences between $d_{\hat{\pi}_{BC}}$ and $d_{\pi^*}$, where the estimator is no longer consistent.

\subsection{Closed-loop Weighted Empirical Risk Minimization}
\label{sec:cwerm}
In our method, called ``Closed-loop Weighted Empirical Risk Minimization'' (CW-ERM),  we seek to debias a policy network from the open-loop performance towards closed-loop performance, making the model rely on features that are robust to closed-loop evaluation. Our method consists of three stages: the training of an identification policy, the use of that policy in closed-loop simulation to identify samples, and the training of a final policy network on a reweighted data distribution. More explicitly:
\vspace{2mm}
\\
\noindent \textbf{Stage 1 (identification policy)}: train a traditional BC policy network in open-loop using ERM, to yield $\hat{\pi}_{\text{ERM}}$.

\noindent \textbf{Stage 2 (closed-loop simulation)}: perform rollouts of the $\hat{\pi}_{\text{ERM}}$ policy in a closed-loop simulator, collect closed-loop metrics and then identify the error set below:
\begin{align}
\label{eqn:error-set}
     E_{\hat{\pi}_{\text{ERM}}} = \{(s_i, a_i)~\text{s.t.}~ {C(s_i, a_i)} > 0 \},
\end{align}

where $\text{s}_i$ is a training data sample, or ``scene'' with a fixed number of timesteps from the training set, $\text{a}_i$ is the action performed during the roll-out and $C(\cdot)$ is a cost such as the number of collisions found during closed-loop rollouts.

\noindent \textbf{Stage 3 (final policy)}: train a new policy using weighted ERM where
the scenes belonging to the error set $E_{\hat{\pi}_{\text{ERM}}}$ are upweighted by a factor $w(\cdot)$, yielding the policy $\hat{\pi}_{\text{CW-ERM}}$:

\begin{equation}
\label{eqn:bc-cw-erm}
\argmin_{\pi \in \Pi} \mathbb{E}_{s \sim d_{\pi^*}, a \sim \pi^*(s)}[w(E_{\hat{\pi}_{\text{ERM}}}, s) \ell(s,a,\pi)]
\end{equation}

As we can see, the CW-ERM policy in Equation~\ref{eqn:bc-cw-erm} is very similar to the original BC policy trained with ERM in Equation~\ref{eqn:bc-erm}, with the key difference of a weighting term based on the error set from closed-loop simulation in Stage 2. In practice, although statistically
equivalent, we upsample scenes by a fixed factor rather than reweighting, as it is known to be more stable and robust~\cite{resampling-outperforms}.

By training a policy using CW-ERM, we expect it to upsample scenes that perform poorly in closed-loop evaluation, making the policy network robust to the covariate shift seen during inference time while unrolling the policy.

We describe the complete CW-ERM training procedure in Algorithm~\ref{alg:cw-erm} and in Figure~\ref{fig:high_level_arch} we show a high-level overview of our method.

\begin{algorithm}[t]
\begin{flushright}
    \begin{algorithmic}
    \textbf{Input:} Training set $\mathcal{D}$ and hyperparameters $K$ (number of epochs) and $w$ (upsampling factor).
    
    \textbf{Stage 1: Identification policy}
    
        1. Train $\hat{\pi}_{\text{ERM}}$ on $\mathcal{D}$ using ERM for $K$ epochs (Equation~\ref{eqn:bc-erm}). 

    \textbf{Stage 2: Closed-loop simulation}

        2. Perform closed-loop simulation of the policy $\hat{\pi}_{\text{ERM}}$ in training scenes;
        
        3. Compute closed-loop evaluation metrics;

        4. Build the error set $E_{\hat{\pi}_{\text{ERM}}}$ of training scenes from closed-loop metrics (Equation~\ref{eqn:error-set}).
    
    \textbf{Stage 3: Final policy}
        
        5. Construct upsampled dataset $\mathcal{D}_{\text{up}}$ by upsampling the error set $E_{\hat{\pi}_{\text{ERM}}}$ by $w$ times;
        
        6. Train final model $\hat{\pi}_{\text{CW-ERM}}$ for $K$ epochs on $\mathcal{D}_{\text{up}}$ via CW-ERM (Equation~\ref{eqn:bc-cw-erm}).
    \end{algorithmic}
    \end{flushright}
    \caption{CW-ERM training procedure}
    \label{alg:cw-erm}
\end{algorithm}

\subsection{Relationship to covariate shift adaptation with density ratio estimation}
One important connection of our method is with covariate shift correction using density ratio estimation~\cite{covshift}. To correct for the covariate shift, the negative log-likelihood is often weighted by the density ratio $r(s)$:

\begin{equation}
\label{eqn:bc-density-ratio}
\argmin_{\pi \in \Pi} \mathbb{E}_{s \sim d_{\pi^*}, a \sim \pi^*(s)}[r(s) \ell(s,a,\pi)]
\end{equation}

where $r(s)$ is defined as the density ratio between test and training distributions:

\begin{equation}
\label{eqn:density-ratio}
r(s) = \frac{p_{\text{test}}(s)}{p_{\text{train}}(s)}
\end{equation}

In practice, $r(s)$ is difficult to compute and is thus estimated. The density ratio will be higher when the sample is more important for the test distribution. In our method (CW-ERM), instead of using the density ratio to weight training samples, we resample the training set based on an estimate of each data point's importance towards good closed-loop behaviours. Like the density ratio, the weighting in our case will also be higher for when the sample is important for the test distribution.

One key characteristic of the importance weighted estimator is that it can be consistent even under covariate shift. We leave, however, the analysis of theoretical properties of our approximation for future work.

\section{Related Work}
Closely related to our work is the ``Learning from Failure'' method~\cite{nam2020learning} (also known as LfF), where the authors train two models at the same time with a similar purpose of mitigating bias. The difference is that in CW-ERM we train models sequentially and we use closed-loop evaluation metrics instead of the loss to upsample, which permits the use of non-differentiable metrics. We also do not use GCE (generalized cross-entropy) to bias the identification model. Our method is simple from a training perspective, but unlike LfF, it does assume the availability of a simulator.

A similar approach that has been successfully applied to computer vision and natural language processing is the \emph{Just Train Twice} (JTT) algorithm~\cite{liu2021just}. JTT similarly first trains an identification model, then trains another model which upweights samples misclassified by the identification model. Although similar in the sense that two models are trained sequentially, in JTT the goal is to deal with worst-group accuracy and not to improve robustness to closed-loop behaviors of a planning model, as in our case.

Several works have attempted to address the covariate shift problem in imitation learning. ChauffeurNet \cite{bansal2018chauffeurnet} and SafetyNet \cite{vitelli2022safetynet} add state perturbation to the training data for improved generalization. Similarly, DAVE-2 \cite{nvidia-end-to-end-2016} used video captured from three different cameras as well as perturbation on the captured images. Another common approach is to supplement training with on-policy data~\cite{ross2011reduction,pan2017agile,prakash2020exploring}, however, in practice, collecting on-policy data for use during training can be extremely expensive and time-consuming.

Similar to our work, Urban Driver~\cite{scheel2022urban} also utilizes a closed-loop simulator, but the simulator is used directly during training to generate unrolls while using BPTT (backpropagation through time). Urban Driver needs a differentiable simulator and does not scale well due to the need for rollouts during training and the memory requirements of BPTT. In contrast, our work takes a much simpler approach where the closed-loop simulator is only used to identify which samples to up-weight and does not require a differentiable simulator, while being able to directly identify scenes that are important for closed-loop evaluation without having to change the training loss to add differentiable collision losses as in Urban Driver~\cite{scheel2022urban}. In our work, any closed-loop metric can be used, with no requirements for differentiability.

\section{Experimental Evaluation}
\label{sec:experimental}
\subsection{Policy network architecture}
Our method is agnostic to model architecture choices. To evaluate our CW-ERM approach, we adopt the recent network architecture of~\cite{vitelli2022safetynet} to represent a strong baseline performance for SDV planning. This model uses a transformer-based~\cite{transformers_all_you_need} architecture with a vectorial input representation~\cite{gao2020vectornet} to create features for each element into vector sets. It consists of a PointNet-based~\cite{qi2016pointnet} module for local processing of vectorized inputs and a global graph using a Transformer encoder for reasoning about interactions with agents and map features. Differently from~\cite{vitelli2022safetynet}, we don't use a safety layer, as we want to evaluate the planner performance without external trajectory fallbacks. For further details, please refer to~\cite{vitelli2022safetynet}.

\subsection{Training}
During training, we found that stopping training of the identification policy before convergence, similar to what was done in JTT~\cite{liu2021just} and LfF~\cite{nam2020learning}, also yielded better results. We limited the capacity of the identification policy by training it until $K$ epochs. The insight is that important biases are learned in early training phases~\cite{nam2020learning} and limiting model capacity can avoid overfitting and avoid depletion~\cite{liu2021just} of the error set used for the training of the final policy. We train the final policy for 40 epochs.

For the ERM baseline, we compare our method against two different experiments where we have the traditional BC trained with ERM with and without perturbations (details can be found in Appendix \ref{append:perturb}).

\subsection{Datasets}
We train and test CW-ERM on a proprietary dataset. Our SDV data is collected in challenging urban missions on San Francisco and Palo Alto roads. This dataset is a collection of driving trajectories from our SDV and surrounding agents, along with recorded HD Maps. Various types of behavioral scenarios in urban driving such as stopping behind a lead vehicle, stopping at intersections, and driving among dense cars, pedestrians, cyclists etc.~are captured. The majority of scenes in our dataset are between 11-13 seconds long, with the longest lasting up to 30 seconds. The total data used during training is 180 hours and we validate and test on 60 hours of driving data each. %

\subsection{Evaluation framework}
We compute the closed-loop evaluation metrics by doing rollouts of the policy in the log-replayed scenes on a simulator~\footnote{Please refer to the \nameref{sec:reproducibility} section for details on the open-sourcing of the simulator and metrics used in this work.}. During the unroll, trajectories are recorded. An evaluation plan composed of a set of metrics and constraints is executed over the recorded trajectories. We count every scene that violated a constraint (e.g., a collision) and then compute the confidence intervals (CIs) for each metric using a Binomial exact posterior estimation with a flat prior, which gives similar results (up to rounding errors) to bootstrapping as recommended in~\cite{rl-statistical-precipice-2021}.

\begin{table*}[ht]
\centering
\captionsetup{justification=centering}
\caption{Experimental results from closed-loop evaluation in simulation. In this table we show a baseline method of behavioral cloning (ERM) with and without perturbations (details can be found in Appendix \ref{append:perturb}) together with the results from single and multi-metric experiments. Lower is better for all metrics.}
\label{tab:model-eval}
\resizebox{\textwidth}{!}{
\begin{tabular}{l| l| l| l l l l}
\hline
\toprule
 \thead{Method} & \thead{Upsampled metric} & \thead{Perturbation} & \thead{Front Collisions} & \thead{Side Collisions} & \thead{Rear Collisions} & \thead{Dist. to ref. traj.}\\
 \midrule
ERM (baseline) & (not applicable) &  & 67 \textcolor{mygray}{(52.8, 85.0)} & 97 \textcolor{mygray}{(79.6, 118.2)} & 114 \textcolor{mygray}{(94.9, 136.8)} & 75 \textcolor{mygray}{(59.9, 93.9)}\\ 
  ERM (baseline) & (not applicable) & \checkmark & 14 \textcolor{mygray}{(8.4, 23.5)} & 55 \textcolor{mygray}{(42.3, 71.6)} & 34 \textcolor{mygray}{(24.4, 47.5)} & 35 \textcolor{mygray}{(25.2, 48.6)}\\ 
  \midrule
  CW-ERM (ours) & Front Collisions & \checkmark & \textbf{9} \textcolor{mygray}{(4.8, 17.1)} & 52 \textcolor{mygray}{(39.7, 68.1)} & 38 \textcolor{mygray}{(27.7, 52.1)} & 39 \textcolor{mygray}{(28.6, 53.2)}\\
  CW-ERM (ours) & Side Collisions & \checkmark & 11 \textcolor{mygray}{(6.2, 19.7)} & 47 \textcolor{mygray}{(35.4, 62.5)} & 31 \textcolor{mygray}{(21.9, 44.0)} & 35 \textcolor{mygray}{(25.2, 48.6)}\\
  CW-ERM (ours) & Dist. to Reference Trajectory & \checkmark & 10 \textcolor{mygray}{(5.5, 18.4)} & 50 \textcolor{mygray}{(37.9, 65.8)} & 40 \textcolor{mygray}{(29.4, 54.4)} & \textbf{28} \textcolor{mygray}{(19.4, 40.4)}\\
  \midrule
  CW-ERM (ours) & Front + Side Collisions & \checkmark & 9 \textcolor{mygray}{(4.8, 17.1)} & 55 \textcolor{mygray}{(42.3, 71.6)} & 39 \textcolor{mygray}{(28.5, 53.2)} & 35 \textcolor{mygray}{(25.2, 48.6)}\\
  CW-ERM (ours) & Front + Side + Rear Collisions & \checkmark & 11 \textcolor{mygray}{(6.2, 19.7)} & 50 \textcolor{mygray}{(37.9, 65.8)} & \textbf{28} \textcolor{mygray}{(19.4, 40.4)} & 37 \textcolor{mygray}{(26.9, 51.0)}\\ 
  CW-ERM (ours) & Front + Side + Dist. to ref. traj. & \checkmark & 10 \textcolor{mygray}{(5.5, 18.4)} & \textbf{46} \textcolor{mygray}{(34.5, 61.3)} & 40 \textcolor{mygray}{(29.4, 54.4)} & 31 \textcolor{mygray}{(21.9, 44.0)}\\ 
\bottomrule
\end{tabular}
}
\end{table*}

\begin{figure*}[h!]
    \centering
\includegraphics[width=\linewidth]{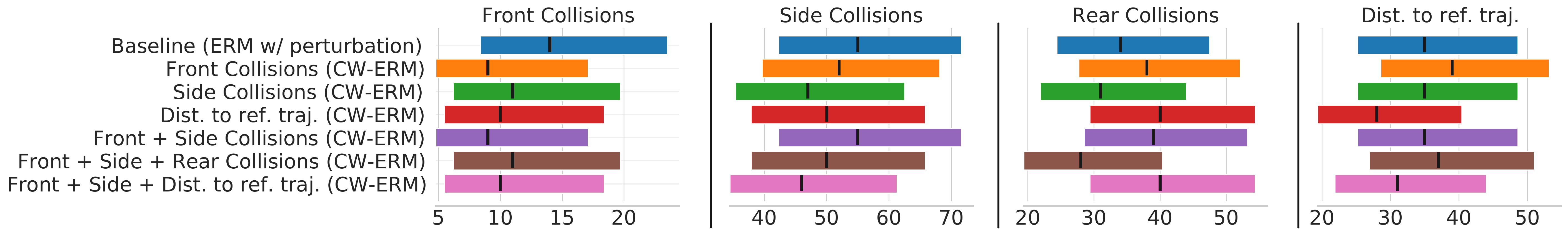}
    \caption{Visual representation of the experimental results from closed-loop evaluation in simulation shown in Table~\ref{tab:model-eval}. Confidence intervals (CIs) were calculated using .95 interval from an exact Binomial posterior with a flat prior. In this plot we only compare against the best baseline (ERM with perturbation).}
    \label{fig:model-eval-CIs}
\end{figure*}

\subsection{Metrics}
Metrics computed in the closed-loop simulator are used to construct the error set~(Equation~\ref{eqn:error-set}). In our evaluation we consider certain important metrics: the number of front collisions, side collisions, rear collisions, and distance from reference trajectory. The distance from reference trajectory considers the entire target trajectory for the current simulated point. A failed scene with respect to this metric is one where the distance of the simulated center of the SDV to the closest point in the target trajectory is farther than four meters. 

In our evaluation, we perform two sets of experiments: \emph{single metric} and \emph{multi metric}. In single metric experiments we construct the error set using only a single metric, while for multi metric we use scenes from multiple metrics together.

\subsection{Results}

\subsubsection{Single Metric}
We show the results from single metric experiments in Table~\ref{tab:model-eval}. We can see that the number of collisions significantly reduced for both side and front collision experiments. We found improvements in the range of $\sim$35\% on the test set for some metrics when compared to the baseline. 

We also found that the largest margin of improvements targeting single metrics in isolation were seen when using single metric based error set, while a balance was achieved when targeting multiple metrics, which suggests a Pareto front of solutions when targeting multiple objectives.

Variance is also lower in some cases when compared to the baseline. We note that while upsampling a certain metric, it shows noticeable improvements in other related metrics. For example, in our single metric experiments, we see that improving side collisions also improve rear collisions. This is evidence that the model is not only getting better at side collisions but also becoming less passive (as indicated by reduction in rear collisions, due to log-replayed agents in simulation that are non-reactive). %

\begin{figure}[ht]
    \centering
\includegraphics[width=\linewidth]{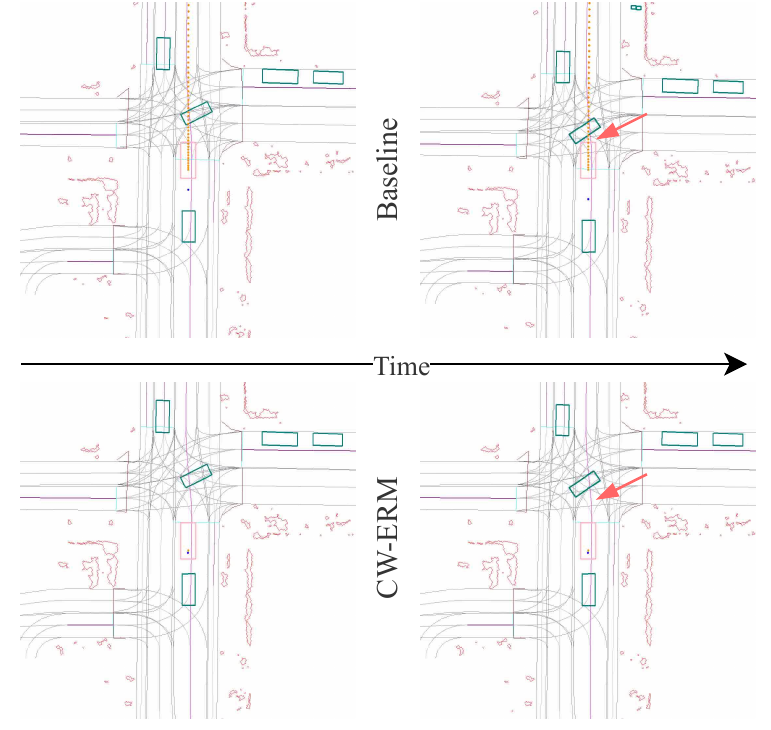}
     \caption{A scene from the test dataset showing the behavior of ego in Baseline (ERM) (with perturbation) and CW-ERM front collisions upsampled. The blue dots are the target trajectory and the yellow dots are the predicted trajectory. The ego is the box in pink and blue-green boxes are other agents. Here, we see that the baseline moves ahead at intersection ignoring the car from the right resulting in a front collision. In contrast, for the CW-ERM policy, it waits at the intersection.}
    \label{fig:scenes_front_coll}
\end{figure}

\begin{figure}[ht]
    \centering
\includegraphics[width=\linewidth]{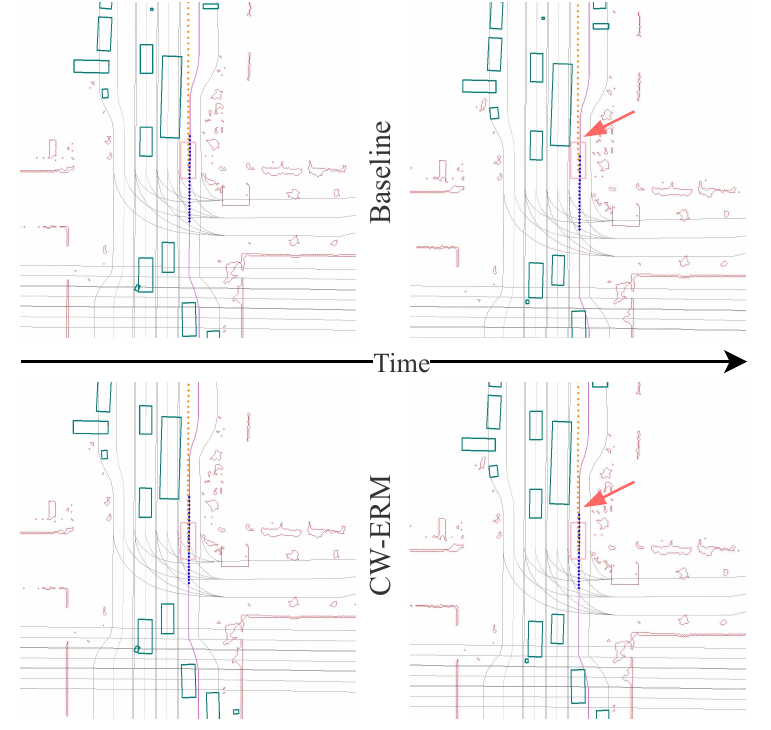}
     \caption{A scene from the test dataset showing the behavior of ego in Baseline (ERM) (with perturbation) and CW-ERM side collisions upsampled. The blue dots are the target trajectory and the yellow dots are the predicted trajectory. The ego is the box in pink and blue-green boxes are other agents. Here, we see that the baseline deviates from target trajectory and collides with the bus ahead. The CW-ERM policy slows down behind the bus and prevents a collision.}
    \label{fig:scenes_side_coll}
\end{figure}

Qualitative results during closed-loop unroll are shown in Figures \ref{fig:scenes_front_coll} and \ref{fig:scenes_side_coll}. Here, we show improved behavior when the scene is upsampled in CW-ERM - once for front collisions Figure \ref{fig:scenes_front_coll} and another for side collisions Figure \ref{fig:scenes_side_coll}. Here, we see a better response in the CW-ERM model which avoids collisions by waiting at intersection and slowing down next to a lead vehicle.
\subsubsection{Multi Metric}
In our multi-metric experiments, we combine two or more metrics - namely $m_{1},m_{2}..m_{N}$ - into a single upsampling experiment. The metrics are equally weighted and hence scenes that fail due to any $m_i$ will be added to the error set. While improvements are noticeable upon combining Front and Side collisions or Front, Side and Distance to the reference trajectory in Table \ref{tab:model-eval}, considerable regression is observed when adding rear collisions. As we can see from the experiments, this is clearly related to the amount of false-positives (FPs) in rear collisions due to the lack of agent reactivity during log playback in the simulator.

\begin{figure}[H]
    \centering
\includegraphics[width=\linewidth]{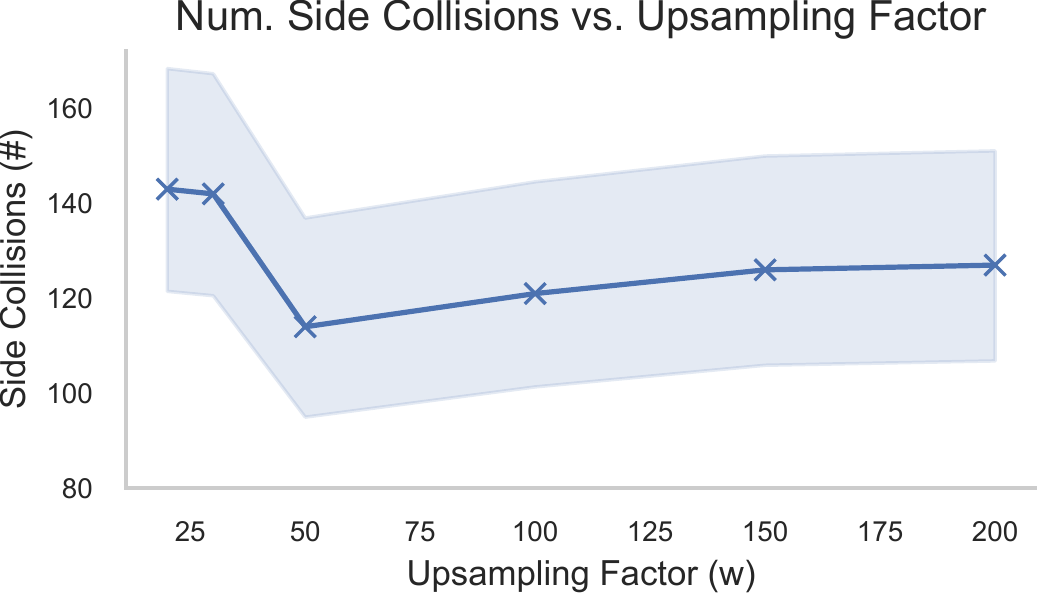}
    \caption{A sample plot showing the effect of upsampling factor on the performance based on single cost upsampling. Here, the scenes are upsampled based on side collisions by factor $w$ along X-axis. The resulting side collisions obtained while evaluating on the validation dataset is obtained on the Y-axis. It is evident from the plot that the performance improves until an upsampling factor of $w = 50$ after which number of side collisions begin to increase.}
    \label{fig:upsample-factors}
\end{figure}
\subsection{Hyper-parameter Tuning / Upsampling Experiments}
We evaluate the performance of the upsampled training set using various identification models $K \in \{10,20\}$ with various upsampling factors $w \in \{10, 20, 30, 50\}$ on the validation dataset.
Single metric upsampling experiments responded better to error set extracted from the training where $K = 10$, while using  $K = 20$ performed better for multi-metric experiments. We find that the size of the resulting upsampled error set influences performance. As seen from Table~\ref{tab:model-eval} and Figure~\ref{fig:upsample-factors}, there exists a limit beyond which performance does not improve during the upsampling experiments. 
A similar observation was noted in JTT~\cite{liu2021just}, where they also found an upsampling factor for which beyond it, worst-group accuracy could not be improved.

\section{Limitations}
Although our method is efficient, easy to implement and showed significant improvements, it also comes with limitations that are important to highlight. In this work, we use a proprietary dataset for evaluation, primarily due to the current lack of available closed-loop evaluation benchmarks. Most public datasets available today are focused on agent prediction tasks and on open-loop metrics (e.g.~\cite{argoverse-dataset-2019,waymo-dataset-2021}). Recently, the closed-loop planning benchmark nuPlan~\cite{nuplan-dataset-2021} was released, but is still under active development and requires a special license for industrial labs to use.

Our method also introduces two new hyper-parameters: $K$ (number of epochs for the identification model) and $w$ (upsampling factor), however, we found the improvement to be robust to different parameterizations of these parameters, similar to past observations~\cite{liu2021just}.

We performed log-replay of agents in simulation, which can produce false positives for the rear collision metric. We leave further analysis and the usage of a reactive simulator as future direction.

We have not yet deployed our policy in a real-world SDV, however, we evaluated it on a closed-loop evaluation framework that is known to be closely representative of real-world performance. Deployment and testing of a policy in a real-world SDV on public roads requires further safety evaluations that we leave as future work.

\section{Reproducibility}
\label{sec:reproducibility}
We make available our closed-loop simulator and the closed-loop metrics used in this work in the following open-source repository: 
\centerline{\url{https://woven-planet.github.io/l5kit}}

\section{Discussion}
Most recent improvements in imitation learning are based on improving the asymptotic performance of algorithms. In this work we showed a different direction that tackles the problem by directly addressing the mismatch between training and inference without requiring an extra human oracle or adding extra complexity during training. Our method is as simple as upsampling scenes by leveraging any existing simulator and training two models, yet it showed that there is still room for significant improvements without having to deal with human-in-the-loop, training rollouts or impacting the policy inference latency. We also described an important potential connection of our method with density ratio estimation for covariate shift correction~\cite{covshift}, which we believe is an exciting future research direction that could provide better theoretical understanding of the improvements seen in our experiments.

\appendix

\subsection{Perturbation}\label{append:perturb}
In our experiments, we employed similar perturbation techniques as in \cite{vitelli2022safetynet}. We randomly perturb the ego's current state to shift from the current trajectory in some of the training examples. To put it more concretely, for the perturbed states, we add a zero-mean Gaussian noise to the ego's current position and heading. We perturb the ego's speed by $av+|b|$ where $v$ is the current speed, $a$ and $b$ are speed multiplier and bias terms generated by a zero-mean Gaussian distribution. Taking the absolute value of the bias term is to ensure that the perturbed speed is always non-negative. Additionally, we perform collision checks for every perturbed state and we do not include any perturbed states which includes collisions.

\subsection{Policy network hyper-parameters}
To train the baselines and our policy we employ a distributed training with a local batch size of 64 for each replica (with an effective batch size of 4096 when using 64 GPU replicas). We use a learning rate of 0.001 that is annealed with cosine schedule during training for 40 epochs (except for the identification models as described in the Section~\ref{sec:experimental}). We also used MAE (mean absolute error) loss and the Adam~\cite{kingma:adam} optimizer with default PyTorch~\cite{NEURIPS2019_9015} parameters for $\beta_1$ and $\beta_2$.

\subsection{Scene distribution}

\begin{figure}[ht!]
\includegraphics[width=\linewidth]{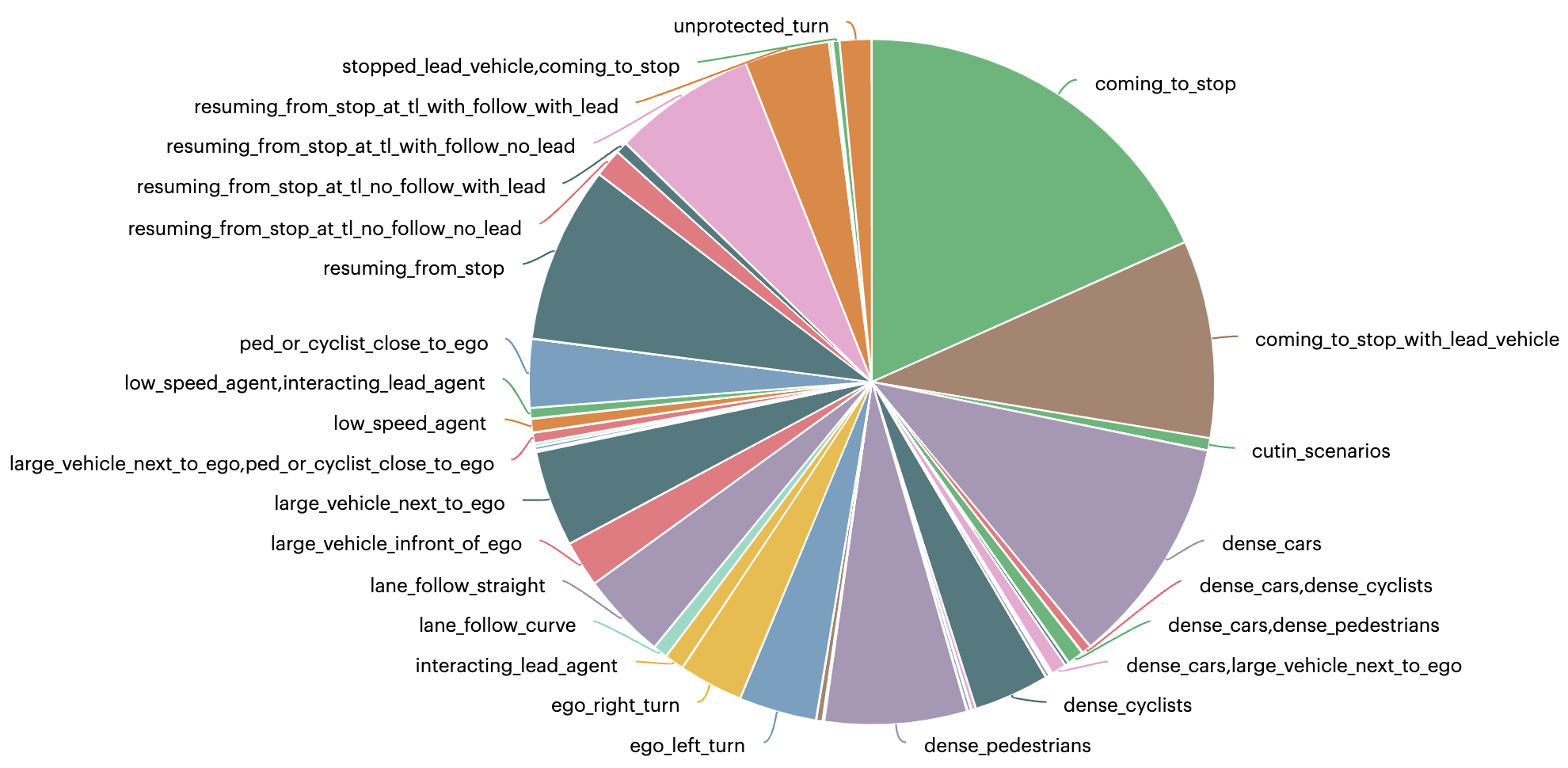}
\caption{Scenario distribution of the training dataset. Scenes can be of single or multiple scenarios - in the case of multiple scenario tags, the scenario names are comma separated.}
  \label{fig:scenario_dist_names}
\end{figure}

We select training, validation and test data subsets to be balanced. The scenarios selected for the datasets are diverse, and a variety of complex urban scenarios have been curated such that it is possible to evaluate closed-loop performance, even on minority scenario groups. In Figure \ref{fig:scenario_dist_names}, a detailed split of the various scenarios is provided.

\section*{ACKNOWLEDGMENT}
We would like to thank Kenta Miyahara, Nobuhiro Ogawa and Ezequiel Castellano for the review of this work and everyone from the UK Research Team and the ML Planning team who supported this work through the ecosystem needed for all experiments, and for the fruitful discussions.

\bibliographystyle{IEEEtran}
\bibliography{robusttrain}

\end{document}